\documentclass[runningheads]{llncs}

\usepackage{eccv}

\usepackage{booktabs}
\usepackage[table]{xcolor}
\usepackage{multirow}
\usepackage{tcolorbox}
\definecolor{skyblue}{RGB}{232, 243, 255}

\newcommand{\Object}{\textsc{Object}}
\newcommand{\Action}{\textsc{Action}}
\newcommand{\Location}{\textsc{Location}}
\newcommand{\ranker}{\texttt{QSRanker}}
\newcommand{\retrieval}{\texttt{QSRetrieval}}
\newcommand{\qsvideo}{\texttt{QSVideo}}
\newcommand{\ourdata}{\texttt{Video-Ranker-35K}}

\DeclareMathOperator*{\argmax}{arg\,max}
\tcbuselibrary{listings,breakable}

\newtcblisting{promptblock}{
  colback=gray!15,
  colframe=gray!60,
  boxrule=0.4pt,
  arc=2pt,
  breakable,
  listing only,
  listing options={
    basicstyle=\scriptsize\ttfamily,
    breaklines=true
  }
}

\usepackage{eccvabbrv}

\usepackage{graphicx}
\usepackage{booktabs}

\usepackage[accsupp]{axessibility}  

\usepackage{hyperref}

\usepackage{orcidlink}

\begin{document}

\title{QSVideo: Query-Conditioned Semantic Temporal Retrieval for Video Understanding} 

\titlerunning{QSVideo}

\author{Wei Ao\orcidlink{0000-0003-1449-936X} \and
Lan Wang\orcidlink{0000-0002-7341-4904} \and
Vishnu Naresh Boddeti\orcidlink{0000-0002-8918-9385}}

\authorrunning{W. Ao et al.}

\institute{Michigan State University, East Lansing MI 48824, USA\\
\email{\{aowei,wanglan3,vishnu\}@msu.edu}}

\maketitle

\begin{abstract}

The performance of vision-language models (VLMs) in video understanding declines with increasing video duration, as video moments unrelated to the query confuse their language components. Multimodal retrieval has emerged as a critical component of video understanding, addressing this challenge by localizing key visual evidence. However, existing multimodal retrieval methods suffer from biased relevance estimation, limited diversity, and temporal collapse. In this paper, we propose QSVideo, a unified framework that systematically addresses relevance, diversity, and temporal modeling in video retrieval. We first introduce a query-conditioned semantic ranker, QSRanker, which reformulates arbitrary questions into retrieval-friendly queries and estimates structured relevance along object, action, and location dimensions. Building upon this, we design QSRetrieval to jointly optimize relevance and diversity for more informative frame selection. Moreover, we propose temporal alignment strategies tailored for both long and streaming videos to improve evidence recall. Extensive experiments on long and streaming video benchmarks demonstrate that QSVideo greatly enhances video VLM performance under strict frame limit constraints. The code is available at \url{https://github.com/human-analysis/QSVideo}.


\end{abstract}

\section{Introduction\label{sec:intro}}

Video understanding involves answering a question using the information from a video. Although vision-language models (VLMs) demonstrate strong video understanding capability~\cite{neurips2023llava, blog2024llavanext, hurst2024gpt4o, chen2024internvl, yao2024minicpm}, their performance often degrades as video duration increases~\cite{wang2024lvbench}. Naively feeding more frames is not a practical solution to this problem, as it does not necessarily improve accuracy while significantly increasing the GPU memory footprint. Humans rarely consult the video uniformly. Instead, they form a goal-driven hypothesis from the question and actively search for supporting visual evidence in the video. They focus on semantically relevant moments, avoid repeatedly examining similar clips, and scan different time periods to gather complementary clues. This behavior naturally balances \emph{relevance}, \emph{diversity}, and \emph{temporal coverage}, motivating a retrieval-centric design for video understanding.

Despite recent progress, question-to-frame retrieval still suffers from three fundamental issues.  First, \textbf{biased relevance}: directly using the original question as a query~\cite{huang2025frag, wang2025seal, buch2025flexible, cvpr2025atp,yu2023self,hu2025selector} can yield unreliable relevance estimates, especially for complex formats such as multiple-choice questions. Second, \textbf{limited diversity}: ranking frames by relevance alone often results in near-duplicate evidence with low information gain~\cite{huang2025frag, buch2025flexible, cvpr2025atp}. Third, \textbf{temporal collapse}: existing methods frequently concentrate selections to a narrow temporal region, although key evidence may appear at distant timestamps in long videos.

\begin{figure}[t]
    \centering
    \includegraphics[width=0.9\linewidth]{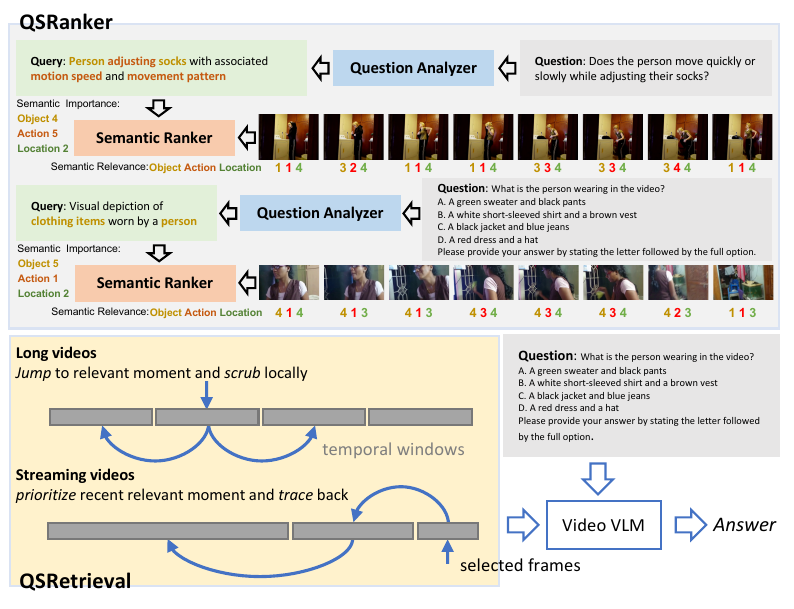}
    \caption{{\qsvideo{}: query-conditioned semantic temporal retrieval for long and streaming videos. \qsvideo{} includes \ranker{}, \retrieval{} and a video VLM. \ranker{} proposes a \emph{Question Analyzer} (LLM) to understand arbitrary questions and transcribe original questions to retrieval-friendly queries. The question analyzer analyzes the questions and estimates the importance of \Object{}, \Action{} and \Location{}. \ranker{} designs a \emph{Semantic Ranker} to score frames in terms of relevance to the query along \Object{}, \Action{} and \Location{} dimensions. \retrieval{} is a novel search framework that balances relevance, diversity, and temporal coverage, inspired by the human video understanding process. We can seamlessly integrate \ranker{}, \retrieval{} with different video VLMs to greatly improve their video understanding ability without scaling up models or expanding frames.} \label{fig:qsvideo}}
\end{figure}

To address these challenges, we propose \qsvideo{}, a query-conditioned retrieval framework for video understanding. \qsvideo{} consists of two key components -- \ranker{} and \retrieval{} as shown in Fig.~\ref{fig:qsvideo}. \qsvideo{} discovers visual evidence by explicitly balancing \emph{relevance}, \emph{diversity}, and \emph{temporal coverage} under a strict frame budget. Motivated by human video processing (question analysis $\rightarrow$ evidence discovery $\rightarrow$ understanding/reasoning), \ranker{} understands and transcribes the questions to queries and evaluates query-to-frame semantic \emph{relevance}. The distance between visual evidences are taken into account to improve the \emph{diversity} of selections. \retrieval{} maximizes the \emph{temporal coverage} for different videos: long video (locate globally relevant anchors and then search locally) and streaming videos (recency-first traversal).

We extensively benchmark \qsvideo{} on challenging long~\cite{wang2024lvbench} and streaming video~\cite{lin2024streaming} benchmarks. 
Under strict frame budgets (8/16/32 frames), \qsvideo{} achieves state-of-the-art performance with an efficient 8B video VLM on StreamingBench~\cite{lin2024streaming} and outperforms the backbone VLM by 6.9-7.1 pp in accuracy on LVBench\cite{wang2024lvbench}, and remains plug-and-play across multiple VLM backbones with consistent gains. 
\qsvideo{} achieves this with only modest computational overhead and can be efficiently accelerated via batched inference, making it well-suited for real-world deployment.
Our contributions are:
\begin{tcolorbox}[width=\linewidth,colback=skyblue,colframe=skyblue,boxsep=0pt,left=3pt,right=3pt]

\begin{enumerate}
    \item \qsvideo{}, a query-conditioned retrieval framework that selectively attends to informative visual evidence and greatly boosts the performance of video VLMs;
    \item \ranker{} to understand and transcribe arbitrary questions into retrieval-friendly queries and estimate structured semantic relevance along \Object{}, \Action{}, and \Location{} dimensions;
    \item \retrieval{} to explicitly balance relevance, diversity, and temporal coverage, with tailored temporal alignment for both long and streaming videos.
\end{enumerate}
\end{tcolorbox}

\section{Related Work \label{sec:related}}

\noindent \textbf{Multimodal Large Language Models (MLLMs)} have bridged the gap between visual perception and linguistic reasoning by integrating pre-trained vision encoders with Large Language Models  \cite{neurips2023llava, zhu2023minigpt4}. Early influential models like LLaVA-1.5 \cite{liu2023llava15} and MiniGPT-4 \cite{zhu2023minigpt4} utilized simple linear layers or MLPs to project image tokens, demonstrating remarkable zero-shot capabilities. Recent iterations, such as LLaVA-NeXT \cite{li2024llavanext} and Insight-V \cite{cvpr2024insight}, have further refined these architectures to support higher-resolution inputs and more complex reasoning tasks. Extending MLLMs from static images to dynamic video content introduces significant challenges, primarily regarding temporal coherence and the token explosion problem, where high frame counts exceed the LLM's context window. Early attempts such as VideoChat \cite{li2023videochat}, Video-ChatGPT \cite{maaz2023videochatgpt}, and Video-LLaVA \cite{lin2023videollava} primarily relied on sparse frame sampling to fit video data into existing image-based pipelines. To address the limitations of long-video comprehension, recent research has moved toward more sophisticated compression and integration strategies. Video-LLaMA \cite{zhang2023videollama} utilizes a Q-Former to bootstrap vision-language alignment, while LLaMA-Vid \cite{li2024llamavid} introduces efficient visual compression to handle extended sequences. Scaling of multimodal pre-training  models like InternVL2.5 \cite{chen2024internvl25}, Qwen2.5-VL \cite{bai2025qwen25vl} also demonstrate superior reasoning across images, videos, and real-time streaming inputs.

\noindent \textbf{Long-form and Streaming Video Understanding}. Recent advancements in long-form and streaming video understanding have introduced diverse innovations to handle extensively long temporal sequences and real-time inputs efficiently \cite{wang2024videollamb, shen2025vgent, shu2025videoxl, kim2025infinipotv, chatterjee2025providellm, zeng2025streamforest, wang2025seal, diko2025rewind, iccv2025openhouse, neurips2025livestar, iccv2025trialtotriumph, iccv2025dynimg, chen2024videollm, wu2024videollm, lin2024streaming, yang2025svbench, qian2024streaming, zhang2025flash, huang2025cvpr}. Early efforts to overcome context window limitations utilized recurrent mechanisms, such as VideoLLaMB~\cite{wang2024videollamb}, which pioneered Recurrent Memory Bridges to propagate information across sequential segments. Streaming video understanding has emerged as a distinct task focusing on online videos, where processing continuously arriving frames allows for real-time interaction in applications like virtual reality and robotics. Unlike offline models, streaming VLMs must simultaneously memorize the past, perceive the current scene, and predict future states~\cite{huang2025cvpr}. To optimize computational overhead, InfiniPot-V~\cite{kim2025infinipotv} enforces a length-independent memory cap through in-place KV cache compression, while ProVideLLM~\cite{chatterjee2025providellm} achieves sub-linear scaling via interleaved summaries. Specialized memory structures have also been proposed, such as Flash-VStream~\cite{zhang2025flash}, which maintains vision feature centers as semantic memory, and StreamForest~\cite{zeng2025streamforest}, which adaptively organizes frames into a Persistent Event Memory Forest based on temporal distance and content similarity. For hour-scale reasoning, Video-XL~\cite{shu2025videoxl} further condenses context into compact Visual Summarization Tokens.

\noindent \textbf{Retrieval and Frame Selection}. Effective information retrieval and selective frame or token sampling are critical for isolating task-relevant evidence from dense video content, ranging from query-agnostic selection to query-cognitive relevance ranking \cite{liu2025bolt, tang2025aks, ye2025tstar, zhang2025flexselect, guo2025logicinframes, zhang2025dvd, ma2025drvideo, pan2026timesearchr, zhou2025reagentv, arnab2025temporalcot, iccv2025flow4agent, iccv2025lvagent, iccv2024vca, neurips2025adavideorag, iclr2026vqos, iclr2026querystream, mrvideo2025, zhang2025llavaVideo, yao2024minicpm, yu2023self, huang2025frag, hu2025selector, neurips2023llava, chen2024internvl, buch2025flexible}. While many models like LLaVA-Video~\cite{zhang2025llavaVideo} and SEAL~\cite{wang2025seal} adopt uniform sampling to maximize temporal diversity, newer methods prioritize query-relevant frames. BOLT~\cite{liu2025bolt} utilizes inverse transform sampling and recursive temporal binning to identify informative frames without model retraining. Logic-in-Frames~\cite{guo2025logicinframes} reformulates this as a semantic-logical search across temporal relations. Alternatively, SeViLA~\cite{yu2023self} and FRAG~\cite{huang2025frag} use VLMs as cross-modal rankers to score frame relevance based on specific queries. However, off-the-shelf VLMs often lack consistency in relevance scores as they are not specifically fine-tuned for cross-modal ranking tasks. To address this, FFS~\cite{buch2025flexible} and TimeSearch-R~\cite{pan2026timesearchr} learn optimized policies to dynamically select frames with self-verification. Furthermore, T*~\cite{ye2025tstar} recasts temporal search as a spatial grid task, and FlexSelect~\cite{zhang2025flexselect} identifies semantically critical visual tokens by analyzing internal attention patterns.

\section{QSVideo: Relevance, Diversity and Temporal Coverage}

We first describe our problem of interest, namely \textbf{query-to-frame retrieval for video understanding}. Then, we propose the query-conditioned semantic ranker (\ranker{}) in \S~\ref{sec:ranker} and introduce automated frame-level video annotation and training in \S~\ref{sec:train}. Finally, we present our solution to our problem of interest, \retrieval{}, in  \S~\ref{sec:retrieval}. \ranker{} followed by \retrieval{} can locate key visual clues to facilitate following video VLMs. \ranker{} and \retrieval{} are agnostic to video VLMs. In this paper, video VLMs enhanced by \ranker{} and \retrieval{} are referred to as \qsvideo{} (Fig.~\ref{fig:qsvideo}).

\noindent\textbf{Problem statement.} Given a query $q$ and a series of video frames $\mathcal{V} = \{v_{i}\}_{i=0}^{T_V}$, the objective is to find a subset $A \subseteq  \mathcal{V}$ such that (1)~$A$ provides the necessary visual information relevant to the query, (2)~$A$ has maximum visual diversity (\ie the minimum visual redundancy), and (3)~$|A| = K$. We formulate it as:
\begin{equation}
\begin{gathered}
\max_{A \subseteq \mathcal{V}}
\Bigg\{
\sum_{v_i \in A} \operatorname{Relevance}(q, v_i),\;
\sum_{\substack{v_i, v_j \in A \\ i \neq j}}
\operatorname{Diversity}(v_i, v_j)
\Bigg\} \\
\text{s.t.}\quad |A| = K, \quad \sum_{i \in A}  \operatorname{T}(v_i) \ge \Gamma
\end{gathered}
\label{eq:retrieval_objective}
\end{equation}
where we further introduce the temporal coverage measure $\sum_{i \in A}  \operatorname{T}(v_i)$ to mitigate \emph{temporal collapse}, where frame selection concentrates on a narrow temporal region. The temporal coverage measure reduces near-duplicate evidence and increases effective information density.
Broader coverage improves evidence recall under a limited frame budget, which is essential for long-range reasoning in long videos, where critical visual clues may be temporally dispersed.

\subsection{QSRanker: Query-Conditioned Semantic Ranker \label{sec:ranker}}

Given a query $q$ and a series of frames uniformly sampled from a video $\mathcal{V} = \{v_i\}_{i=1}^{T_V}$, the goal is to \emph{score} frames based on their relevance with respect to the query $q$. Frames with higher relevance scores provide more visual evidence required to answer the query. Existing works, \eg SeViLA~\cite{yu2023self}, FRAG~\cite{huang2025frag}, Frame Selector~\cite{hu2025selector}, Qwen3-VL-Reranker~\cite{qwen3vlembedding}, prompt pretrained VLMs to generate ``yes'' or ``no'' for each query-frame pairs. This approach has three key drawbacks: (1)~they do not analyze and leverage \emph{hints} from the query, (2)~a single binary score cannot represent the semantic relationship between a query and frames,
and (3) they adopt original questions as queries, likely leading to biased relevance because of wrong options from multiple-choice questions. SEAL~\cite{wang2025seal} introduces three different models to detect scenes, objects, and actions, and ignores the semantic understanding ability of VLMs.

We introduce \textbf{Q}uery-conditioned \textbf{S}emantic \textbf{Ranker}, dubbed \ranker{}, that integrates a \emph{question analyzer} and a \emph{semantic ranker}. The question analyzer transcribes arbitrary questions to retrieval-friendly queries while exploring hints from questions. The semantic ranker scores query-frame pairs by their \Object{}-, \Action{}-, and \Location{}-level semantic relevance. We describe each module in detail below.

\noindent\textbf{Question Analyzer.} In real-world applications, questions may appear in diverse formats (e.g., open-ended or multiple-choice) and span heterogeneous tasks such as recognition, reasoning, and temporal grounding. Directly using the original question as the retrieval query can lead to \emph{incorrect relevance estimation}. For instance, in multiple-choice settings, distractor options introduce misleading semantic cues; if a ranker attends to these incorrect options, it may assign high relevance scores to irrelevant frames. In contrast, when humans are asked a question, they first identify the key evidence required to answer it. Inspired by this process, we introduce a question analyzer to interpret arbitrary questions and transcribe them into retrieval-friendly queries. Our question analyzer extracts the core intent of the query while filtering out noisy or task-specific artifacts, producing a clean and retrieval-oriented representation.

The relative importance of \Object{}, \Action{}, \Location{} evidence varies significantly across queries. For example, some queries emphasize specific entities, while others focus on dynamic events or physical locations. As a result, effective frame retrieval requires adapting the contribution of different semantic aspects to the query. Our question analyzer captures this query-dependent preference and yields important weights along  \Object{}-, \Action{}-, and \Location{}-dimensions. 

In this paper, we use a Qwen3-4B~\cite{qwen3} as the question analyzer $\operatorname{W}$:
\begin{equation}
    (q, w_o, w_a, w_l) = \operatorname{W}(\hat{q})
    \label{eq:weight}
\end{equation}
where $\hat{q}$ is the original question and $q$ is the transcribed question, \ie the query. The question analyzer assigns importance weights $w_o, w_a, w_l$ to \Object{}, \Action{}, and \Location{}, respectively. The importance weights are discrete values from $\{1,2,3,4,5\}$, with higher values indicating greater importance.
They are determined solely by the query and reflect how critical each type of semantic evidence is to answering the query.
These weights are used to aggregate semantic relevance scores in the following scoring, enabling query-conditioned frame retrieval.

\noindent\textbf{Semantic Ranker.} Conditioning on the query, relevant evidence may arise from the presence attributes of specific entities (\Object{}), dynamic events and temporal changes (\Action{}), or spatial relations (\Location{}). To model such semantic relevance, we apply a vision–language model that takes a query-frame pair as input and predicts structured semantic relevance scores along the \Object{}, \Action{}, and \Location{} dimensions:
\begin{equation}
    (o, a, l) = \operatorname{R}_\theta(q, v)
    \label{eq:ranker}
\end{equation}
where $\operatorname{R}_\theta$ denotes the VLM ranker with trainable parameter $\theta$, and $(o, a, l)\in\{1,2,3,4,5\}$ are the predicted relevance scores corresponding to \Object{}, \Action{}, and \Location{} respectively. These scores provide an interpretable representation of how each frame supports the query and serve as the basis for frame ranking and selection. Rather than relying on explicit semantic annotations or handcrafted rules, we model relevance implicitly by learning how different semantic aspects of a frame align with the query. In this paper, the semantic ranker is a lightweight pretrained VLM Qwen3-VL-4B-Instruct~\cite{qwen3vl} and is finetuned on a frame-level ranking video dataset.

\noindent\textbf{Scoring.} Given the semantic relevance scores $\operatorname{R}_\theta(q, v) = (o, a, l)$ and the query-adaptive weights
$(w_o, w_a, w_l)$, we compute a unified ranking score for each frame:
\begin{equation}
    \operatorname{Score}(q, v)
    =  \tilde w_o \cdot o + \tilde w_a \cdot a + \tilde w_l \cdot l
    \label{eq:score}
\end{equation}
where $(\tilde w_o, \tilde w_a, \tilde w_l)$ are normalized importance weights via $\operatorname{Softmax}$. 

The proposed \ranker{} decouples question understanding from visual grounding, leverages the question analyzer (LLM) only for estimating evidence importance and generating retrieval-friendly query, and allows the semantic ranker to focus on learning fine-grained visual relevance. 

\subsection{Training the Semantic Ranker \label{sec:train}}

To align a pretrained VLM with the semantic ranking (Eq.~\ref{eq:ranker}), we need to collect video data, annotate videos at the frame level, and fine-tune the pretrained VLM. The resulting semantic ranker can understand the query, the visual clues, their semantic relationship, and how to score them along \Object{}, \Action{}, and \Location{} dimensions, explicitly enforcing the correct ranking for query-frame pairs.

\noindent \textbf{Video Data.} 
We construct a fine-tuning dataset, \ourdata{}, by randomly sampling 35K videos from LLaVA-Video-178K~\cite{zhang2025llavaVideo}. 
For both single-round and multi-round conversations, we retain only one question per video, resulting in 35K video–query pairs. 
Among them, 28,848 videos are shorter than 30 seconds, and 6,154 are shorter than 60 seconds. \ourdata{} covers three video QA formats: open-ended (75.9\%), captioning (14.5\%), and multiple-choice (9.6\%). The videos originate from diverse sources within LLaVA-Video-178K, including YouTube (78.9\%)~\cite{xue2022advancing,wang2023internvid,zhu2023languagebind}, Charades (9.5\%)~\cite{sigurdsson2016hollywood}, YouCook2 (6.0\%)~\cite{zhou2017youcookii}, NextQA (2.9\%)~\cite{xiao2021next}, ActivityNet (2.4\%)~\cite{caba2015activitynet}, and Ego4D (0.3\%)~\cite{grauman2022ego4d}. 

\noindent \textbf{Semantic Annotation.} For each (video, question) instance, we first prompt the question analyzer to transcribe the question ($\hat{q}$) to a query ($q$). Then, we uniformly sample frames $\mathcal{V}=\{v_i\}_{i=1}^{T_V}$ and form $T_V$ query-frame pairs $\{(q, v_i)\}_{i=1}^{T_V}$ where $T_V=8$. The semantic annotation task is to obtain the frame-level semantic relevance scores $\{(o_i, a_i, l_i)\}_{i=1}^{T_V}$. 

Manual annotation of fine-grained semantic scores is prohibitively expensive. The state-of-the-art large-scale VLMs (\eg GPT-5~\cite{singh2025openai}, Gemini 3~\cite{gemini3}, Qwen3-VL~\cite{qwen3vl}, \etc) can serve as automatic annotators to generate semantic relevance labels. In this paper, we leverage the strong off-the-shelf VLM Qwen-VL-30B~\cite{qwen3vl} to produce semantic supervision at scale. We prompt Qwen-VL-30B to generate discrete semantic relevance scores, \ie $o, a, l \in \{1,2,3,4,5\}$ for all query-frame pairs. Therefore, we build the triplet dataset for supervised fine-tuning, $\mathcal{D}_{\text{SFT}}=\{(q, v_i, y_i)_{i=1}^{T_V}\}$ where $y_i=(o_i,a_i,l_i)$.  

\noindent\textbf{Supervised Fine-Tuning.} We fine-tune the semantic ranker $\operatorname{R}_\theta(q, v)$ via supervised fine-tuning (SFT)~\cite{ouyang2022training} on the annotated~\ourdata{}.
\begin{equation}
\mathcal{L}_{\text{SFT}}(\theta)
=  
\mathbb{E}_{(q,v,y)\sim\mathcal{D}_{\text{SFT}}}
\left[
\ell_{\text{CE}}(\hat o, o)+
\ell_{\text{CE}}(\hat a, a)+
\ell_{\text{CE}}(\hat l, l)
\right]
\label{eq:sft}
\end{equation}
where $y=(o,a,l)$ and $\ell_{\text{CE}}$ is the Cross-Entropy loss over the 5 discrete score tokens. After SFT training, the semantic ranker can generate aligned structured relevance scores along \Object{}, \Action{}, and \Location{} for individual query-frame pairs.

\subsection{QSRetrieval: Relevance–Diversity Temporal Retrieval \label{sec:retrieval}}

\textbf{Visual Distance.} To prevent redundant visual appearance, we need to measure the visual distance of selected frames. To that end, we obtain global visual representations using the vision encoder of the semantic ranker, \ie $g=f(v)$. Specifically, we aggregate embeddings of all vision tokens via mean pooling to produce a global embedding. We define their visual distance as the Euclidean (L2) distance, \ie $d_{ij} = \left\| g_i - g_j \right\|_2$. We empirically observe L2 distance is more sensitive to visual change, compared to cosine similarity, which measures only angular difference and discards magnitude information. L2 distance captures both directional and norm variations in the embedding space. L2 distance better reflects subtle appearance and layout changes between frames, making it more suitable for redundancy reduction.

\noindent \textbf{Relevance–Diversity Retrieval.} To balance query-to-frame relevance and visual diversity in Eq.~\ref{eq:retrieval_objective}, we iteratively select a new frame from the complement set $A^{c} = \mathcal{V} \setminus A$:
\begin{equation}
    v^\ast = \argmax_{v \in A^c} \Bigg\{ \lambda \cdot \operatorname{Score}(q, v) + (1 - \lambda) \min_{v^\prime \in A} \left\| f(v) - f(v^\prime) \right\|_2 \Bigg\}
    \label{eq:relevance_diversity}
\end{equation}
where $A$ is the current selection and $\mathcal{V}$ is the whole video. We greedily search for $v$ so that it maximizes the query-to-frame relevance score and the visual distance with respect to selected frames. 

\noindent\textbf{Temporal Windowing.} Eq.~\ref{eq:relevance_diversity} guides the retrieval to attend to both relevance and diversity. However, it neglects temporal coverage introduced by Eq.~\ref{eq:retrieval_objective}. To encourage temporal coverage, we divide the video frame set 
$\mathcal{V}$ into $K$ disjoint temporal windows 
$\{\mathcal{V}_i\}_{i=1}^{K}$ according to their timestamps, such that
\begin{equation}
\mathcal{V} = \bigcup_{i=1}^{K} \mathcal{V}_i, 
\quad
\mathcal{V}_i \cap \mathcal{V}_j = \emptyset \ \text{for } i \neq j
\label{eq:partition}
\end{equation}
After partitioning the whole video into $K$ temporal windows, we restrict the selection within each window to ensure temporal coverage. Specifically, for each window $\mathcal{V}_i$, we select at most one frame by
\begin{equation}
v_i^\ast
=
\argmax_{v \in \mathcal{V}_i}
\Bigg\{
\lambda \cdot \operatorname{Score}(q, v)
+
(1 - \lambda)
\min_{v^\prime \in A}
\left\| f(v) - f(v^\prime) \right\|_2
\Bigg\}
\label{eq:temporal_relevance_diversity}
\end{equation}

\noindent\textbf{Temporal-Aware Retrieval.} While Eq.~\ref{eq:temporal_relevance_diversity} enforces window-level coverage, the retrieval across temporal windows should adapt to different videos. In this paper, we focus on two challenging tasks: \emph{long} video understanding and \emph{streaming} video understanding. 

Motivated by how humans search for evidence in videos, we couple temporal windowing with an adaptive traversal strategy.
When answering questions on \emph{long videos}, people typically jump to the most related moment in memory and then search locally around it; for \emph{streaming videos}, people instead prioritize recent context and progressively trace back in time.
We instantiate these two behaviors with different window-ordering and reference-updating rules while keeping the same relevance--diversity scoring.

\begin{itemize}
    \item \textbf{Long Videos.} Let $\mathcal{I}=\{1,\dots,K\}$ be temporal window indices and maintain a set of available windows $\mathcal{U}\subseteq \mathcal{I}$ (initialized as $\mathcal{U}=\mathcal{I}$). At iteration $t$, we first choose a \emph{global anchor} from the remaining windows:
    \begin{equation}
    a^{(t)}=\argmax_{v\in \bigcup_{i\in \mathcal{U}}\mathcal{V}_i} \operatorname{Score}(q,v)
    \end{equation}
    We denote the corresponding window index of $a^{(t)}$ by $i^{(t)}$.  Then, we search locally within a temporal radius $r$ around $i^{(t)}$:
    \begin{equation}
    \mathcal{N}^{(t)}=\{\, i\in \mathcal{U}\mid |i-i^{(t)}|\le r \,\}
    \end{equation}
    We apply Eq.~\ref{eq:temporal_relevance_diversity} to search over neighboring windows $i\in \mathcal{N}^{(t)}$ and use the anchor as the reference to compute visual distance: 
    \begin{equation}
        v_i^{\ast}=\argmax_{v\in \mathcal{V}_i}
        \Big\{
        \lambda\,\operatorname{Score}(q,v) + (1-\lambda)\,\|f(v)-f(a^{(t)})\|_2
        \Big\}
        \label{eq:long_video}
    \end{equation}
    After this local expansion, we mask out visited windows by updating $\mathcal{U}\leftarrow \mathcal{U}\setminus \mathcal{N}^{(t)}$ and repeat the process
    until $K$ windows are selected. This iterative \emph{anchor-and-expand} scheme naturally recovers evidence that may appear in multiple distant temporal regions.
    
    \item \textbf{Streaming Videos.} For streaming videos, we traverse windows from recent to past. Starting from the most recent window, we first select
    \begin{equation}
    v_K^\ast=\argmax_{v\in \mathcal{V}_K}\operatorname{Score}(q,v)
    \end{equation}
    We then progressively move to earlier windows $i=K-1,\dots,1$, using the previous selection as the reference, and select
    \begin{equation}
        v_i^{\ast}=\argmax_{v\in \mathcal{V}_i}
        \Big\{
        \lambda\,\operatorname{Score}(q,v) +
        (1-\lambda)\,\|f(v)-f(v_{i+1}^{\ast})\|_2
        \Big\}
    \end{equation}
    This recency-first traversal explicitly prioritizes near-term evidence by construction, while the distance term discourages selecting visually redundant frames across adjacent windows.
    
\end{itemize}
\section{Experiments}

We fine-tune the proposed semantic ranker using SFT on~\ourdata{}. Training is conducted on 8$\times$ NVIDIA A6000 GPUs (48GB VRAM each) with DeepSpeed ZeRO-2~\cite{rajbhandari2020zero} and bfloat16 precision. We adopt parameter-efficient tuning via LoRA~\cite{hu2022lora} (rank=32, $\alpha$=64). The global batch size is 128 with cosine learning rate decay (base LR $1\times10^{-4}$, warmup ratio 3\%). SFT takes approximately 12 hours for one epoch. During SFT, we freeze the LLM backbone and fine-tune the vision tower and merger modules, because the semantic ranker mainly requires improving the visual–language alignment between query and frames, which is determined by the vision tower and merger modules. If we fine-tune the LLM on the query-frame dataset~\ourdata{}, it may affect its strong language understanding ability. In our experiments, \qsvideo{} adopts MiniCPM-o 2.6~\cite{yao2024minicpm} as the backbone video VLM.

\subsection{Long Video Understanding \label{sec:lvbench}}

\begin{table*}[t]
\centering
\caption{Experiments on long video understanding (LVBench). LVBench has six types of tasks: entity recognition (ER), event understanding (EU), key information retrieval (KIR), temporal grounding (TG), reasoning (Rea) and summarization (Sum).  \label{tab:lvbench}}
\resizebox{0.9\linewidth}{!}{
\begin{tabular}{lccccccccc}
\toprule
\textbf{Model} & \textbf{LLM} & \textbf{Frames} & \textbf{Overall} & \textbf{ER} & \textbf{EU} & \textbf{KIR} & \textbf{TG} & \textbf{Rea} & \textbf{Sum} \\

\midrule
\rowcolor{gray!10}
Deep Video Discovery~\cite{zhang2025dvd} & - & - & 74.2 & 73.4 & 73.3 & 80.4 & 72.3 & 70.6 & 74.1 \\
\rowcolor{gray!10}
Seed1.5-VL-Thinking~\cite{guo2025seedvl} & 200B & - & 64.6 & 65.4 & 63.4 & 68.0 & 53.6 & 63.7 & 46.6 \\
\rowcolor{gray!10}
AdaReTaKe~\cite{wang2025adaretake} & 72B & $\leq$1024 & 53.3 & 53.0 & 50.7 & 62.2 & 45.5 & 54.7 & 37.9 \\

\midrule
GLM-4V-Plus~\cite{glm4v} & - & 30 & 38.3 & 39.9 & 35.8 & 34.8 & 37.7 & 40.0 & 32.8 \\
MiniCPM-o 2.6~\cite{yao2024minicpm} & 8B & 16 & 38.9 & 37.1 & 38.2 & 37.5 & 32.7 & 46.3 & 41.4 \\
InternVL2-40B~\cite{chen2024far} & 34B & 16 & 39.6 & 37.4 & 39.7 & 43.4 & 31.4 & 42.5 & 41.4 \\
Qwen2-VL-72B~\cite{wang2024qwen2vl} & 72B & 48 & 41.3 & 38.0 & 41.1 & 38.3 & 41.4 & 46.5 & 46.6 \\
TimeMarker~\cite{chen2024timemarker} & 8B & $\leq$128 & 41.3 & 42.8 & 39.1 & 34.9 & 38.7 & 38.2 & 48.8 \\
MiniCPM-o 2.6~\cite{yao2024minicpm} & 8B & 32 & 42.0 & 41.4 & 40.7 & 39.2 & 35.5 & 48.3 & 37.9 \\
MiniCPM-o 2.6~\cite{yao2024minicpm} & 8B & 64 & 42.3 & 43.0 & 42.5 & 39.9 & 39.1 & 46.8 & 31.0 \\
InternVL2.5-78B~\cite{chen2024expanding} & 72B & 16 & 43.6 & 43.8 & 42.0 & 42.1 & 36.8 & 51.0 & 37.9 \\
\rowcolor{orange!20}
\qsvideo{} (Ours) & 8B & 16 & 45.8 & 47.3 & 41.7 & 49.8 & 35.5 & 50.8 & 32.8 \\
SEAL~\cite{wang2025seal} & 34B & 16 & 45.9 & 47.9 & 41.3 & 51.5 & 32.3 & 43.3 & 39.7 \\
GLM-4V-Plus~\cite{glm4v} & - & $\leq$300 & 48.7 & 46.2 & 47.8 & 54.1 & \textbf{42.7} & 46.5 & 37.9 \\
GPT-4o~\cite{hurst2024gpt4o} & - & 60 & 48.9 & 48.9 & \textbf{49.5} & 48.1 & 40.9 & 50.3 & \textbf{50.0} \\
\rowcolor{orange!20}
\qsvideo{} (Ours) & 8B & 32 & \textbf{49.1} & \textbf{51.1} & 45.6 & \textbf{54.6} & 39.6 & \textbf{53.2} & 32.8 \\ 

\bottomrule
\end{tabular}
}
\end{table*}

\noindent The long video understanding benchmark LVBench~\cite{wang2024lvbench} has 103 long videos and around 15 QAs per video, resulting in 1,549 QAs in total. The average video duration is \textbf{67 minutes}. LVBench comprehensively includes 6 types of tasks: entity recognition (677), event understanding (647), key information retrieval (291), reasoning (201), temporal grounding (220), and summarization (58). We benchmark \qsvideo{} under different frame budgets (16 frames and 32 frames) and compare it with the state-of-the-art video VLMs, as shown in Table~\ref{tab:lvbench}.

\noindent \textbf{Effective Visual Evidence Discovery.} It is challenging to discover relevant visual clues from long videos because visual clues are sparsely distributed. \qsvideo{} achieves an overall accuracy of \textbf{49.1}\%, demonstrating that \retrieval{} can effectively identify informative frames for answering diverse questions. \qsvideo{} achieves the best performance on several tasks that require accurate evidence discovery, including \emph{entity recognition}, \emph{key information retrieval}, and \emph{reasoning}. These tasks heavily depend on locating relevant moments in long videos. The improvements demonstrate that the proposed query-aware semantic ranking can effectively capture structured visual evidence aligned with the query. 

\noindent \textbf{Boosting Video VLM Performance.} The backbone VLM (MiniCPM-o 2.6) reports accuracy of $38.9\%$, $42.0\%$ and $42.3\%$ using 16 frames, 32 frames and 64 frames. \qsvideo{} greatly improves accuracy by $+6.9\%$ and $+7.1\%$ using 16 frames and 32 frames. It indicates that explicitly modeling query-to-frame retrieval can significantly improve the performance of video VLMs.

\noindent \textbf{Efficient Long Video Understanding.} \qsvideo{} achieves competitive performance using only an existing 8B VLM and 16/32 frames, while several competing approaches rely on substantially larger models ($\geq 34$B) or significantly more frames ($\gg 32$). Our results suggest that locating key visual evidence can be more effective than naively scaling up model size or frame budgets for long video understanding, which requires a dramatic increase in the VRAM footprint.

\begin{figure}[t]
    \centering
    \includegraphics[width=0.95\linewidth]{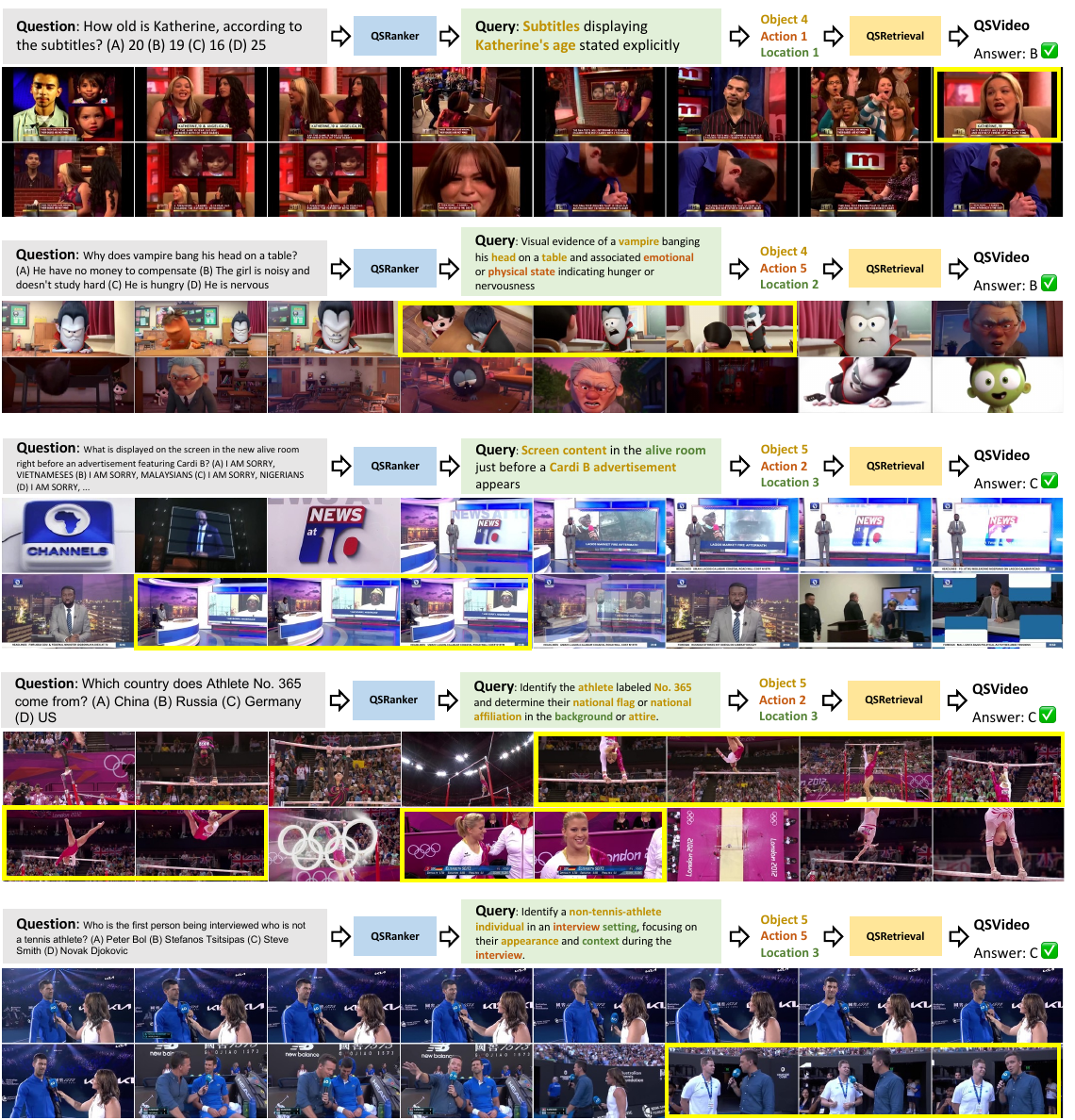}
    \caption{Visualization of QSVideo on LVBench. The number of selected frames is 16. We highlight frames including important supportive visual evidence. \label{fig:lvbench}}
\end{figure}

\noindent \textbf{Visualization.}  We visualize results of \qsvideo{} on LVBench in Fig.~\ref{fig:lvbench}. In Fig.~\ref{fig:lvbench}, we show (1) the original questions, (2) the retrieval-friendly queries generated by \ranker{}, (3) the importance weights of \Object{}, \Action{} and \Location{} generated by \ranker{}, (4) the selected frames retrieved by \retrieval{} and (5) the answers obtained by \qsvideo{}. From Fig.~\ref{fig:lvbench}, we observe that (1) \ranker{} effectively transcribes the original questions and correctly measures the importance of \Object{}, \Action{} and \Location{}; (2) \retrieval{} effectively locates the visual evidence; (3) \qsvideo{} effectively uses these visual evidence and obtains the correct answers.  For example, the question, ``Which country does Athlete No. 365 come from? (A) China (B) Russia (C) Germany (D) US'', \qsvideo{} first finds the Athlete No. 365 and then figures out the nationality from the screen content.

\subsection{Streaming Video Understanding \label{sec:streambench}}

Unlike \emph{offline} long video understanding, we cannot access to the whole video in streaming video understanding.  Because real-world applications require \emph{online} understanding of live videos, \eg AR glasses, live streaming, and autonomous driving. In streaming videos, frames arrive sequentially and future content is not accessible. So, online video VLMs process streaming videos and respond to temporally aligned real-time queries~\cite{chen2024videollm,wu2024videollm,lin2024streaming,qian2024streaming}. 

\begin{table*}[t]
\centering
\scriptsize
\setlength{\tabcolsep}{3pt}
\caption{Experiments on streaming video understanding (StreamingBench). StreamingBench real-time benchmark includes ten types of tasks: object perception (OP), causal reasoning (CR), clips summarize (CS), attribute perception (ATP), event understanding (EU), text-rich understanding (TR), prospective reasoning (PR), spatial understanding (SU), action perception (ACP), counting (CT). \label{tab:stream}}
\resizebox{\linewidth}{!}{
\begin{tabular}{l c c c c c c c c c c c c c}
\toprule
\textbf{Model} & \textbf{LLM} & \textbf{Frames} & \textbf{Overall}
& \textbf{OP} & \textbf{CR} & \textbf{CS} & \textbf{ATP} & \textbf{EU} & \textbf{TR} & \textbf{PR} & \textbf{SU} & \textbf{ACP} & \textbf{CT} \\
\midrule

MiniCPM-o 2.6~\cite{yao2024minicpm} & 8B & \multirow{5}{*}{8} & 64.44 & 67.30 & 71.09 & 74.45 & 71.24 & 71.43 & 64.49 & 55.56 & 59.35 & 62.89 & 35.75 \\
MiniCPM-V 2.6~\cite{yao2024minicpm} & 8B &  & 66.36 & 70.30 & 65.62 & 79.50 & 76.80 & 67.08 & 67.29 & 66.67 & 55.28 & 59.49 & \textbf{45.60} \\
InternVL2~\cite{chen2024far} & 8B &  & 68.52 & 76.02 & 60.94 & 72.24 & 82.35 & 68.94 & 70.72 & 72.22 & 66.67 & 60.62 & 41.97 \\
LLaVA-OV~\cite{li2024llavaov} & 7B &  & 72.12 & 79.29 & 72.66 & 82.97 & 82.35 & 72.67 & 69.16 & 75.00 & 68.29 & 69.12 & 37.31 \\
\rowcolor{orange!20}
\qsvideo{} (Ours) & 8B &  & \textbf{77.68} & \textbf{83.11} & \textbf{79.69} & \textbf{86.12} & \textbf{83.66} & \textbf{80.12} & \textbf{83.49} & \textbf{76.85} & \textbf{72.76} & \textbf{74.22} & 44.04 \\

\midrule

VILA-1.5~\cite{liu2024nvila} & 8B & 14 & 61.54 & 71.12 & 57.81 & 74.68 & 72.22 & 70.81 & 62.62 & 51.85 & 51.22 & 60.34 & 18.65 \\
MiniCPM-o 2.6~\cite{yao2024minicpm} & 8B & 16 & 65.12 & 69.21 & 77.34 & 76.66 & 72.22 & 72.67 & 64.17 & 56.48 & 58.54 & 62.89 & 31.61 \\
MiniCPM-V 2.6~\cite{yao2024minicpm} & 8B & 16 & 70.24 & 75.20 & 71.09 & 83.28 & 80.39 & 68.94 & 68.85 & 70.37 & 59.76 & 66.29 & 46.63 \\
InternVL2~\cite{chen2024far} & 8B & 16 & 70.52 & 75.20 & 64.84 & 75.08 & \textbf{84.97} & 75.16 & 72.90 & 73.15 & 65.85 & 64.59 & 42.49 \\
VITA-1.5~\cite{fu2025vita} & 7B & 16 & 70.88 & 77.66 & 82.81 & 82.33 & 79.34 & 72.67 & 71.34 & 67.59 & 62.20 & 69.12 & 32.12 \\
LLaVA-OV~\cite{li2024llavaov} & 7B & 16 & 73.76 & 79.56 & 79.69 & 82.65 & 83.66 & 73.29 & 74.77 & 72.22 & 69.11 & 69.97 & 40.93 \\
Claude 3.5~\cite{anthropic2024claude35sonnet} & - & 20 & 74.04 & 82.45 & 73.77 & 82.43 & 82.40 & 76.39 & \textbf{85.56} & 61.68 & 60.73 & 67.88 & \textbf{47.62} \\

QueryStream~\cite{iclr2026querystream} & 7B & 1fps & 74.04 & 82.11 & \textbf{83.59} & 78.23 & 82.69 & 75.47 & 80.06 & \textbf{79.63} & 63.01 & 67.90 & 42.55 \\

InfiniPot-V~\cite{kim2025infinipotv} & 7B & 0.5fps & 76.4 & - & - & - & - & - & - & - & - & - & - \\

StreamForest~\cite{zeng2025streamforest} & 7B & 1fps & 77.26 & 83.11 & 82.81 & 82.65 & 84.26 & 77.50 & 78.19 & 76.85 & 69.11 & 75.64 & 54.40 \\

\rowcolor{orange!20}
\qsvideo{} (Ours) & 8B & 16 & \textbf{79.36} & \textbf{84.47} & 82.81 & \textbf{88.96} & 84.31 & \textbf{81.99} & 85.05 & 75.00 & \textbf{74.39} & \textbf{77.34} & 44.56 \\

\bottomrule
\end{tabular}
}
\end{table*}

\begin{table*}[t]
\centering
\scriptsize
\setlength{\tabcolsep}{3pt}
\caption{Seamless plug-and-play over different video VLMs. We integrate \qsvideo{} with different vodeo VLMs (MiniCPM-V 2.6~\cite{yao2024minicpm}, InternVL2~\cite{chen2024far} and LLaVA-OV~\cite{li2024llavaov}) under different frames. \label{tab:plug_and_play}}
\resizebox{\linewidth}{!}{
\begin{tabular}{l c c c c c c c c c c c c c c}
\toprule
\textbf{Model} & \textbf{Frames}
& \textbf{OP} & \textbf{CR} & \textbf{CS} & \textbf{ATP} & \textbf{EU} & \textbf{TR} & \textbf{PR} & \textbf{SU} & \textbf{ACP} & \textbf{CT} & \textbf{Overall} & $\Delta$ \\

\midrule

MiniCPM-V & \multirow{2}{*}{8}
& 70.30 & 65.62 & 79.50 & 76.80 & 67.08 & 67.29 & 66.67 & 55.28 & 59.49 & 45.60 & 66.36 & \multirow{2}{*}{+4.60$\uparrow$} \\
+\qsvideo{} &
& 76.57 & 64.06 & 82.97 & 81.05 & 73.29 & 74.14 & 73.15 & 59.35 & 65.72 & 45.08 & 70.96 & \\
MiniCPM-V & \multirow{2}{*}{16}
& 75.20 & 71.09 & 83.28 & 80.39 & 68.94 & 68.85 & 70.37 & 59.76 & 66.29 & 46.63 & 70.24 & \multirow{2}{*}{+1.60$\uparrow$} \\
+\qsvideo{} &
& 77.66 & 69.53 & 85.80 & 81.37 & 73.29 & 73.52 & 74.07 & 62.60 & 65.72 & 41.97 & 71.84 & \\
MiniCPM-V & \multirow{2}{*}{32}
& 77.66 & 68.75 & 82.02 & 80.72 & 68.94 & 73.52 & 75.00 & 59.35 & 64.59 & 45.60 & 70.80 & \multirow{2}{*}{+2.16$\uparrow$} \\
+\qsvideo{} &
& 78.75 & 71.09 & 85.49 & 83.01 & 72.67 & 76.32 & 75.93 & 60.57 & 67.71 & 45.08 & 72.96 & \\
\midrule

InternVL2 & \multirow{2}{*}{8}
& 76.02 & 60.94 & 72.24 & 82.35 & 68.94 & 70.72 & 72.22 & 66.67 & 60.62 & 41.97 & 68.52 & \multirow{2}{*}{+2.76$\uparrow$} \\
+\qsvideo{}  &
& 78.75 & 67.19 & 75.39 & 85.95 & 76.40 & 71.96 & 70.37 & 64.63 & 65.44 & 44.04 & 71.28 & \\
InternVL2 & \multirow{2}{*}{16}
& 75.20 & 64.84 & 75.08 & 84.97 & 75.16 & 72.90 & 73.15 & 65.85 & 64.59 & 42.49 & 70.52 & \multirow{2}{*}{+1.72$\uparrow$} \\
+\qsvideo{}  &
& 77.93 & 64.84 & 78.86 & 85.95 & 74.53 & 75.08 & 75.00 & 66.26 & 66.57 & 43.52 & 72.24 & \\
InternVL2 & \multirow{2}{*}{32}
& 76.29 & 67.19 & 80.44 & 85.29 & 72.67 & 74.77 & 75.00 & 64.63 & 62.89 & 45.08 & 71.52 & \multirow{2}{*}{+1.40$\uparrow$} \\
+\qsvideo{}  &
& 79.02 & 68.75 & 80.44 & 86.60 & 72.67 & 76.01 & 74.07 & 66.26 & 66.86 & 44.04 & 72.92 & \\

\midrule

LLaVA-OV & \multirow{2}{*}{8}
& 79.29 & 72.66 & 82.97 & 82.35 & 72.67 & 69.16 & 75.00 & 68.29 & 69.12 & 37.31 & 72.12 & \multirow{2}{*}{+2.36$\uparrow$} \\
+\qsvideo{} &
& 80.38 & 75.00 & 85.80 & 83.66 & 73.29 & 76.01 & 75.93 & 70.33 & 71.95 & 37.31 & 74.48 & \\
LLaVA-OV & \multirow{2}{*}{16}
& 79.56 & 79.69 & 82.65 & 83.66 & 73.29 & 74.77 & 72.22 & 69.11 & 69.97 & 40.93 & 73.76 & \multirow{2}{*}{+0.92$\uparrow$} \\
+\qsvideo{} &
& 82.29 & 74.22 & 86.44 & 83.33 & 75.16 & 76.32 & 73.15 & 67.48 & 71.10 & 40.93 & 74.68 & \\
LLaVA-OV & \multirow{2}{*}{32}
& 80.65 & 78.12 & 82.33 & 84.31 & 70.19 & 75.08 & 75.00 & 66.67 & 69.97 & 40.41 & 73.56 & \multirow{2}{*}{+0.84$\uparrow$} \\
+\qsvideo{} &
& 81.74 & 78.12 & 84.23 & 81.70 & 69.57 & 77.88 & 74.07 & 70.33 & 69.97 & 41.97 & 74.40 & \\
\bottomrule
\end{tabular}
}
\end{table*}

We benchmark \qsvideo{} on real-time video understanding using StreamingBench~\cite{lin2024streaming}. StreamingBench real-time benchmark extensively comprises ten tasks: object perception, causal reasoning, clip summarization, attribute perception, event understanding, text-rich understanding, prospective reasoning, spatial understanding, action perception, and counting. 
It includes 2,500 QAs from 499 unique videos,  which provides a comprehensive evaluation of a VLM's working and episodic memory under streaming settings.  The query can be proposed at any time $t$. We follow the standard evaluation of StreamingBench to set the context time to $(t-60, t)$, where $t$ is the query time. We compare \qsvideo{} with existing video VLMs under the same frame budgets, 8 frames and 14--16 frames, as shown in Table~\ref{tab:stream}. We also compare with three recent methods that sample frames by 0.5--1 fps.

\noindent \textbf{Retrieval Matters in Short Contexts.} One might assume that retrieval is less important in streaming understanding because the context window is already short (60 seconds), and uniform sampling should be sufficient. However, Table~\ref{tab:stream} shows that \qsvideo{} still improves performance by $+11.32\%$ and $14.24\%$ using 8 frames and 16 frames compared to the backbone VLM, MiniCPM-o 2.6. Enhanced by \retrieval{}, \qsvideo{} improves the overall accuracy from $72.12\%$ (LLaVA-OV) to $77.68\%$ with 8 frames, and further achieves SoTA accuracy of $79.36\%$ with only 16 frames. It proves that \retrieval{} can effectively discover informative visual evidence within a short temporal context, leading to much better online video understanding.

\noindent \textbf{Better Integrates Visual Clues.} We note the improvements are particularly evident on tasks requiring reasoning over multiple visual cues, such as clip summarize, event understanding, and prospective reasoning (Table~\ref{tab:stream}). For instance, under 16 frames, \qsvideo{} achieves $81.99\%$ on event understanding, significantly higher than most video VLMs. These tasks require integrating observations that may appear at different moments within the recent context window. By selecting frames that are semantically aligned with the query, \qsvideo{} preserves complementary pieces of evidence across time, enabling the model to better connect temporally separated clues when answering streaming queries.

\noindent \textbf{Plug-and-Play over video VLMs.} We further apply \qsvideo{} to different online video VLMs, including MiniCPM-V 2.6~\cite{yao2024minicpm}, InternVL2~\cite{chen2024far}, and LLaVA-OneVision~\cite{li2024llavaov}, under 8-, 16-, and 32-frame settings. Table~\ref{tab:plug_and_play} shows consistent overall improvements across all backbones. It demonstrates that \qsvideo{} is model-agnostic and works in a plug-and-play manner. The experiments prove that discovering informative visual clues can facilitate different video VLMs, especially under limited frames. At 8 frames, \qsvideo{} improves MiniCPM-V, InternVL2, and LLaVA-OV by $+4.60\%$, $+2.76\%$, and $+2.36\%$, respectively.

\subsection{Efficiency}

\begin{table}[t]
\centering
\caption{End-to-end latency. \label{tab:latency}}
\resizebox{0.8\linewidth}{!}{
\begin{tabular}{lcccccc}
\toprule
Model & LLM & Candidate & Selected & Overall (\%) & Latency (s) $\downarrow$ & Throughput $\uparrow$ \\
\midrule
SEAL~\cite{wang2025seal} & 34B & 2045 & 16 & 45.9 & 1702.24 & 1.2fps\\
AdaReTaKe~\cite{wang2025adaretake} & 72B & 1024 & - & 53.3 & 178.15 & 5.8fps \\
\qsvideo{}  & 8B & 2048 & 32 & 49.1 & 91.17 & 22.5fps \\
\qsvideo{}  & 8B & 2048 & 16 & 45.8 & 88.73 & 23.1fps \\
\bottomrule
\end{tabular}
}
\end{table}

\begin{table}[t]
\centering
\caption{Breakdown of latency. \label{tab:breakdown}}
\small
\renewcommand{\arraystretch}{1.2}
\setlength{\tabcolsep}{4pt}
\resizebox{\linewidth}{!}{
\begin{tabular}{l|ccc|cc|cc}
\toprule
\multirow{2}{*}{Acceleration}
& \multicolumn{3}{c|}{QSRanker}
& \multicolumn{2}{c|}{QSRetrieval} & \multicolumn{2}{c}{VLM Inference} \\
 & Visual Distance & Question Analyzer & Semantic Ranker &  16 Frames & 32 Frames &  16 Frames & 32 Frames \\
\midrule
Naive &  16.81s &  1.15s &  608.40s &  0.87s & 1.10s & 8.71s &  11.05s \\
\midrule
+Batch Inference &  16.78s &  1.32s &  167.07s &  0.88s & 1.05s &  8.68s &  10.94s \\
\midrule
+Data Parallel &   16.78s &   1.15s &  61.32s &  0.89s &  1.05s &  8.59s &   10.87s \\
\bottomrule
\end{tabular}
}
\end{table}

We evaluate the end-to-end latency of \qsvideo{} on a long video (video key: Za2Z\_JRxCuk) from LVBench with a duration of 34 minutes, as shown in Table~\ref{tab:latency}. \qsvideo{} uniformly samples 2048 candidate frames from the original video and selects 16 or 32 frames. SEAL~\cite{wang2025seal} samples frames at 1fps resulting in 2045 candidate frames and selects 16 frames, while AdaReTake~\cite{wang2025adaretake} compresses 1024 frames. \qsvideo{} is faster than SEAL by $18.7\times$--$19.2\times$ and faster than AdaReTake by $2.0\times$. We list the detailed latency of components in Table~\ref{tab:breakdown}. \qsvideo{} can be easily accelerated from two aspects: (i) lower end-to-end latency due to lightweight VLMs and (ii) higher throughput due to the parallelizable semantic ranker. 

\section{Ablation Studies}

\subsection{Ablation Study of $\lambda$}

We benchmark the effect of $\lambda$ on LVBench under different frame budgets (Table~\ref{tab:ablation_lambda}). We observe: (i) the performance is relatively insensitive to $\lambda$, which supports ``plug-and-play'' feature, (ii) there is a strong positive relationship between $\lambda$ and the performance (Pearson correlation coefficient $r=0.88$) under 32 frames, which proves that $\lambda=0.9$ effectively balances relevance and diversity. 

\begin{table}[t]
\centering
\caption{Ablation study of $\lambda$. \label{tab:ablation_lambda}}
\resizebox{0.5\linewidth}{!}{
\begin{tabular}{ccccccccc}
\toprule
Frames & $\lambda$ & Overall & ER & EU & KIR & TG & Rea & Sum \\

\midrule

  & 0.1 & 46.1 & 46.7 & 44.2 & 49.8 & 37.3 & 50.2 & 32.8 \\
  
  & 0.3 & 45.6 & 47.3 & 43.3 & 49.1 & 33.6 & 48.3 & 31.0 \\
  
16  & 0.5 & 46.0 & 47.6 & 42.3 & 51.2 & 35.9 & 50.7 & 31.0 \\
  
  & 0.7 & 45.6 & 47.0 & 41.7 & 50.9 & 35.0 & 50.2 & 31.0 \\ 
  
  & 0.9 & 45.8 & 47.3 & 41.7 & 49.8 & 35.5 & 50.8  & 32.8 \\

\midrule

  & 0.1 & 48.2 & 49.0 & 45.1 & 53.6 & 38.2 & 51.7 & 34.5 \\
  
  & 0.3 & 48.3 & 51.1 & 44.4 & 54.3 & 36.4 & 51.2 & 31.0 \\
  
32 & 0.5 & 48.2 & 51.3 & 44.5 & 54.0 & 35.5 & 52.2 & 29.3 \\ 
 
 & 0.7 & 48.7 & 51.3 & 45.0 & 55.0 & 39.1 & 51.7 & 32.8 \\
 
 & 0.9 & 49.1 & 51.1 & 45.6 & 54.6 & 39.6 & 53.2 & 32.8 \\

\bottomrule
\end{tabular}
}
\end{table}

\subsection{Ablation Study of \ranker{}}

We benchmark \ranker{} on LVBench. 1549 QAs have ``time\_reference'' annotations, which are used as ground-truth for ranking evaluation. \ranker{} first uniformly samples 2048 frames and then scores frames. Recall@K measures whether at least one of the top-K ranked frames falls within the ``time\_reference'' interval. The baseline is a holistic ranker that only generates a single score. The baseline reports 17.38\% Recall@1, 26.81\% Recall@5, 32.43\% Recall@10,  46.77\% Recall@50, 52.78\% Recall@100, while \ranker{} reports 17.51\% Recall@1, 27.20\% Recall@5, 33.20\% Recall@10, 50.65\% Recall@50, 58.20\% Recall@100.  These results demonstrate that \ranker{} effectively improves evidence ranking.

\section{Conclusion}

In this paper, we present \qsvideo{}, a multimodal retrieval framework to improve video understanding. \qsvideo{} designs \ranker{} to analyze and transcribe arbitrary questions to retrieval-friendly queries, and score frames along \Object{}-, \Action{}-, and \Location{}-dimensions. \qsvideo{} further proposes  \retrieval{} to discover visual evidence by jointly considering relevance, redundancy reduction, and temporal coverage. On LVBench, \qsvideo{} achieves the best trade-off between model size (8B), the number of frames (32 frames), and QA accuracy ($49.1\%$). On StreamingBench, \qsvideo{} reports very competitive QA accuracy $77.68\%$ and $79.36\%$ by only using 8 and 16 frames. We integrate \qsvideo{} with different video VLMs and observe consistent improvements, especially for limited frames.

\section*{Acknowledgements}
This work was supported in part by the National Science Foundation (award \#2500983). Any opinions, findings, and conclusions or recommendations expressed in this material are those of the authors and do not necessarily reflect the views of NSF.

%
%
\clearpage
\bibliographystyle{splncs04}
\bibliography{main}

\clearpage

\section{Prompts}

\noindent \textbf{Question Analyzer} is a component of \ranker{} that analyzes arbitrary questions, transcribes the original questions to retrieval-friendly queries and generates the importance of \Object{}, \Action{} and \Location{}. In our paper, question analyzer is a lightweight LLM (4B). We use the following prompt:
\begin{promptblock}
<|im_start|>user
<Instruct>: You are assisting a video frame retrieval system.\n\n
Given a Question about a video, determine what types of visual evidence are needed to determine the correct answer.\n\n
Step 1: Rewrite the question into a concise evidence-needed Query (<=30 words).\n
- Do NOT answer the question.\n
- Do NOT copy multiple-choice options.\n 
- Do NOT invent specific details.\n
- Describe only the type of visual evidence needed.\n\n
Step 2: Assign importance weights (1-5) for three dimensions.\n
IMPORTANT: Assess each dimension independently. Judge Object, Action, and Location 
on their own merit for answering the question-do not let one assessment influence another.\n
1) Object (specific entities)\n
2) Action (events or motions)\n
3) Location (environment/background)\n\n
Weight scale:\n
1 = not important\n
2 = slightly important\n
3 = moderately important\n
4 = very important\n
5 = critically important\n\n
Output exactly two lines:\n
Line 1: <Query>\n
Line 2: <object_weight> <action_weight> <location_weight>\n\n
Question:\n
{question}\n\n<|im_end|>
\end{promptblock}

\noindent\textbf{Semantic Ranker} is a component of \ranker{}, that scores the semantic relevance between a query and a frame, along the \Object{}-, \Action{}- and \Location{}-dimensions. In our paper, semantic ranker is a lightweight VLM (4B) fine-tuned on our fine-grained annotated dataset \ourdata{}. We use the following prompt:
\begin{promptblock}
<|im_start|>user
<image>\n
Given a Frame extracted from a video and a Query about the video: '{query}', score the Frame on three dimensions needed to answer the Query:\n
1) Object (specific entities)\n
2) Action (events or motions)\n
3) Location (environment/background)\n\n
Score scale (integers only):\n
1 = not visible / not supported\n
2 = weak hint (ambiguous)\n
3 = partially supported (some clear evidence)\n
4 = strongly supported (clear evidence)\n
5 = decisive match (unambiguous)\n\n
IMPORTANT RULES:\n
- Score based ONLY on what is directly visible in this single frame.\n
- Score each dimension independently. Do not let one dimension influence another.\n
- If the evidence for a dimension is NOT clearly visible, score it 1.\n
- Use 5 only when the frame provides decisive evidence for that dimension.\n\n
Output exactly three integers separated by spaces in this order:\n
<object_score> <action_score> <location_score>\n
Do not output anything else.<|im_end|>
\end{promptblock}

\noindent \textbf{Video QA.} After \qsvideo{} discovers the supportive visual evidence, we can feed the selected frames to any video VLM, \eg MiniCPM-o 2.6~\cite{yao2024minicpm}, MiniCPM-V 2.6~\cite{yao2024minicpm}, InternVL2~\cite{chen2024far}, and LLaVA-OV~\cite{li2024llavaov} \etc. In our experiments, we use the standard prompt provided by the benchmark if available (StreamingBench~\cite{lin2024streaming}). Otherwise, we use the following prompt: 
\begin{promptblock}
<|im_start|>user
<Instruct>: <Frame1>: <image>\n
<Frame2>: <image>\n
<Frame3>: <image>\n
...
You are an advanced video question-answering AI assistant. 
You have been provided with some frames from the video and a multiple-choice question related to the video. 
Your task is to carefully analyze the video and provide the best answer to question, choosing from the four options provided. 
Respond with only the letter (A, B, C, or D) of the correct option.
Question: {question}\n\n
The best option is:<|im_end|>
\end{promptblock} 
 
\section{Video-Ranker-35K}

\begin{figure}
    \centering
    \includegraphics[width=\linewidth]{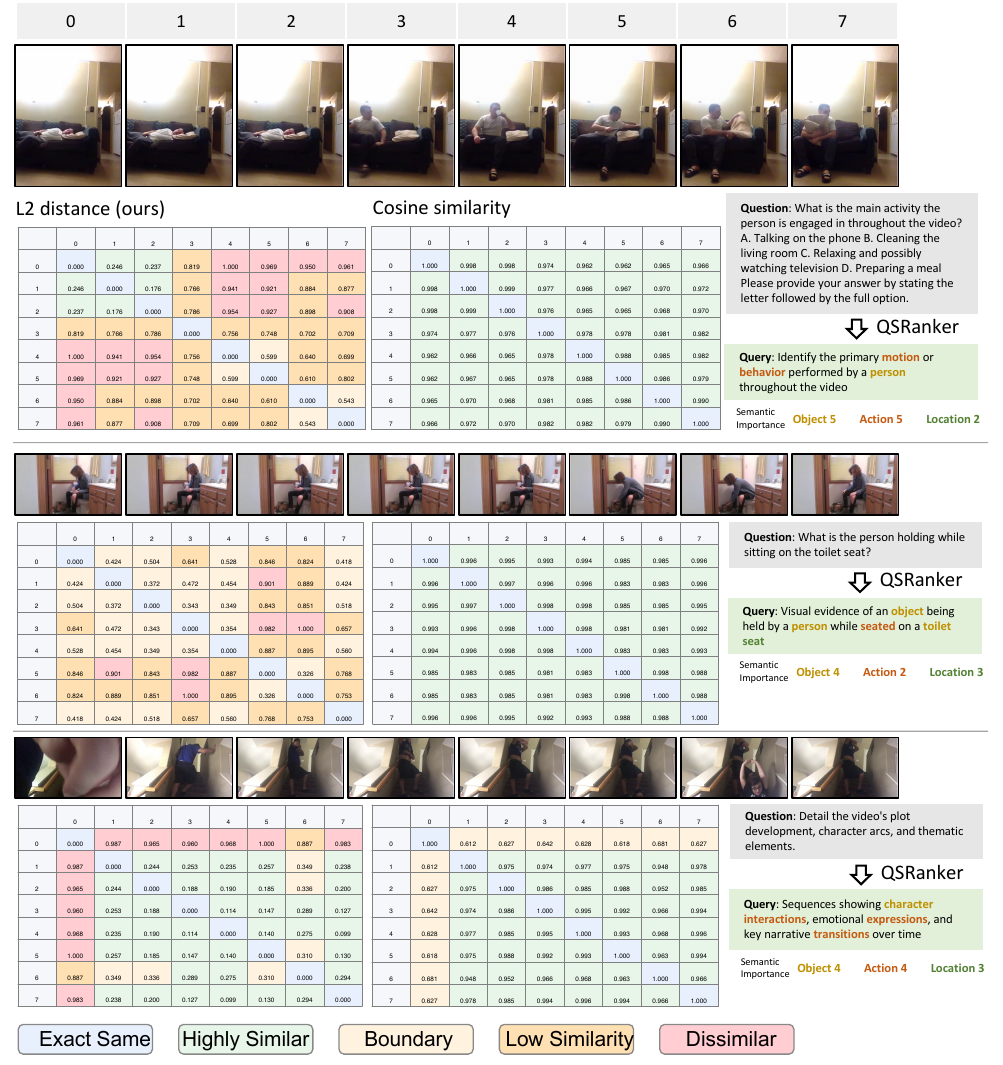}
    \caption{Visualization of Video-Ranker-35K. Comparison between different visual distance matrices: L2 distance (ours) versus cosine similarity. \label{fig:video_ranker_35k}}
\end{figure}

\noindent \textbf{Video-Ranker-35K.}
To fine-tune \ranker{}, we randomly sample 35K videos from LLaVA-Video-178K~\cite{zhang2025llavaVideo} and construct query-frame relevance dataset. Fig.~\ref{fig:video_ranker_35k} shows query-video instances in \ourdata{}. 

\noindent \textbf{Visual Distance.} To prevent redundant visual appearance in the selected frames, \retrieval{} uses the pooling of all visual tokens as the global visual representation of frames and maximizes their L2 distance. Existing work uses cosine similarity to measure visual distance between frames. We empirically observe that L2 distance is better to capture the difference of visual contents between frames, as shown in Fig.~\ref{fig:video_ranker_35k}, leading to better retrieval performance in \retrieval{}.

\end{document}